\documentclass{article}

     \usepackage[preprint]{neurips_2022}

\usepackage[utf8]{inputenc} 
\usepackage[T1]{fontenc}    
\usepackage{hyperref}       
\usepackage{url}            
\usepackage{booktabs}       
\usepackage{amsfonts}       
\usepackage{nicefrac}       
\usepackage{microtype}      
\usepackage{xcolor}         
\usepackage{graphicx}
\usepackage{subfig}

\title{Pruning Early Exit Networks}

\author{
  Alperen Görmez and Erdem Koyuncu\\
  Department of Electrical and Computer Engineering\\
  University of Illinois Chicago\\
  Chicago, IL\\
  \texttt{\{agorme2, ekoyuncu\}@uic.edu}
}

\begin{document}
\maketitle

\begin{abstract}
  Deep learning models that perform well often have high computational costs. In this paper, we combine two approaches that try to reduce the computational cost while keeping the model performance high: pruning and early exit networks. We evaluate two approaches of pruning early exit networks: (1) pruning the entire network at once, (2) pruning the base network and additional linear classifiers in an ordered fashion. Experimental results show that pruning the entire network at once is a better strategy in general. However, at high accuracy rates, the two approaches have a similar performance, which implies that the processes of pruning and early exit can be separated without loss of optimality.
\end{abstract}

\section{Introduction}
Modern deep learning models require large amount of computation to achieve good performance \cite{jhoward, gpt3}. This is undesirable for scenarios in which the model will run on power limited devices such as mobile phones, drones and IoT devices.

The problem of reducing the amount of computation while keeping the model performance high have been tackled from many different angles. Knowledge distillation approach aims to train a smaller model to mimic the output of a larger model \cite{distillation}. Quantization assigns less bits to the model weights to reduce the computational cost \cite{quantization, deepcompression}. Adaptive computation methods propose skipping some parts of the model based on the input \cite{spatial, adaptive, skipnet}. Pruning methods remove redundant weights in the network \cite{pruning, pruningsurvey, rethinkingpruning}. Early exit networks have additional exit points so that not every sample will have to go through the entire network \cite{e2cm, shallowdeep, branchynet}. 

Most early exit networks have two components: an off-the-shelf base network, and additional small classifiers \cite{e2cm, shallowdeep, branchynet}. Until now, no work has been done to prune early exit networks. This paper aims to fill this gap. In particular, we evaluate the performance of two strategies. In the first strategy, we prune and train the additional internal classifiers that enable early exit jointly with the base network. In the second strategy, we first prune and train the base network, and then the classifiers. The significance of the second strategy is that its optimality implies one can separate the processes of early exit and pruning without loss of optimality.


\section{Experiment}
We compare the following two approaches using a ResNet-56 and CIFAR-10 \cite{resnet, cifar}:
\begin{enumerate}
    \item Multiple linear layers are added to a ResNet-56 (the resulting network is a Shallow-Deep Network \cite{shallowdeep}) and the entire model is pruned.
    \item A ResNet-56 is pruned, multiple linear layers are added to the resulting network and then these linear layers are pruned.
\end{enumerate}

The number, locations, and the architectures of the additional linear layers are as described in \cite{shallowdeep}. For both approaches, pruning is followed by fine tuning. The procedure of pruning and fine tuning is repeated 20 times. At each pruning phase, $10\%$ of the weights are pruned using global unstructured $l_1$ norm based pruning. At each fine tuning phase, the model is trained for 10 epochs. Accuracy vs. FLOPs graphs are obtained using different confidence thresholds and applying time sharing as in \cite{e2cm, shallowdeep}.

\begin{figure}[h]

  \subfloat[Sparsity rates of the weights of each layer after 20 iterations of pruning and fine tuning.]{\centerline{\includegraphics[width=0.46\linewidth]{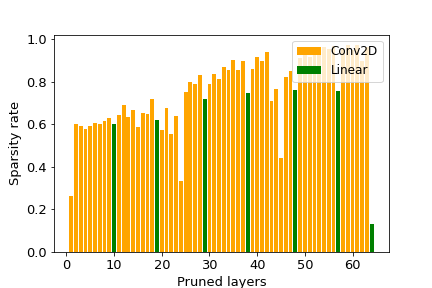}}\label{approach1:a}}
  
  \centerline{ \subfloat[The exit performances after 20 iterations of pruning and fine tuning. Each point corresponds to the scenario where all samples exit from that exit location. Black curve represents the performance before any pruning.]{\includegraphics[width=0.46\linewidth]{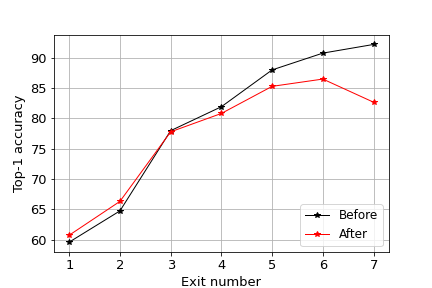}\label{approach1:b}}
  
  \hspace{0.5cm}
  
  \subfloat[Accuracy-FLOPs plot. Samples exit from different exit points due to confidence thresholds. Black curve represents the performance before any pruning.]{\includegraphics[width=0.46\linewidth]{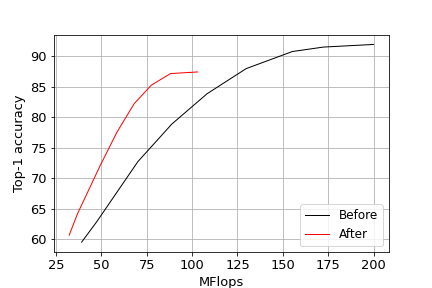}\label{approach1:c}}}
  
  \caption{The sparsity rates and exit performances for Approach 1.}
  \label{fig:approach1}
\end{figure}
\section{Results}
\subsection{Approach 1}
The sparsity rate of each layer's weights and the exit performances after 20 iterations of pruning and fine tuning are shown in Fig.~\ref{fig:approach1}. We can make the following observations:

\begin{enumerate}
    \item Deeper layers are pruned slightly more than earlier layers, and final exit is pruned much less than earlier exits as seen from Fig.~\ref{approach1:a}.
    
    \item Despite extensive pruning, exit performances at earlier exits are comparable with the unpruned baseline, and even better at the first and second exit points as seen from Fig.~\ref{approach1:b}. This suggests pruning can improve model performance.
    
    \item Performance at the last exit is most likely hurt by the joint training of all linear classifiers.
    
    \item Pruning reduces the computational cost up to around $20\%$ without loss of performance. Best accuracy obtained with pruning is around $4\%$ less than the unpruned baseline, but the computational cost is reduced to half as seen from Fig.~\ref{approach1:c}.
\end{enumerate}

\subsection{Approach 2}
\begin{figure}[h]

  \centerline{\subfloat[Sparsity rates for the base network's layers after 20 iterations of pruning and fine tuning.]{\includegraphics[width=0.46\linewidth]{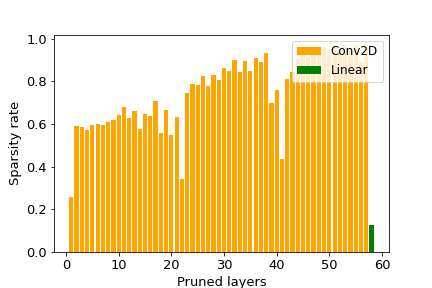}\label{approach2:a}}
  
  \hspace{0.5cm}
  
  \subfloat[Sparsity rates for the additional classifiers after 20 iterations of pruning and fine tuning.]{\includegraphics[width=0.46\linewidth]{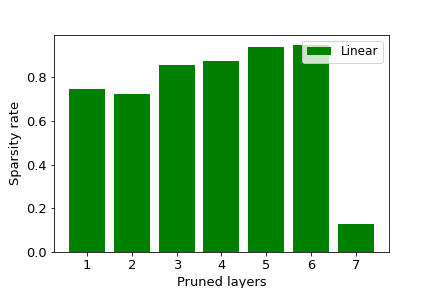}\label{approach2:b}}}
  
  \centerline{\subfloat[The exit performances after 20 iterations of pruning and fine tuning. Each point corresponds to the scenario where all samples exit from that exit location. Black curve represents the performance before any pruning.]{\includegraphics[width=0.46\linewidth]{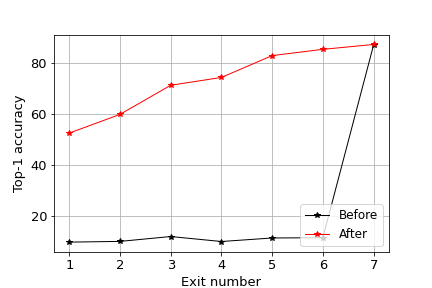}\label{approach2:c}}
  
  \hspace{0.5cm}
  
  \subfloat[Accuracy-FLOPs plot. Samples exit from different exit points due to confidence thresholds. Black curve represents the performance before any pruning.]{\includegraphics[width=0.46\linewidth]{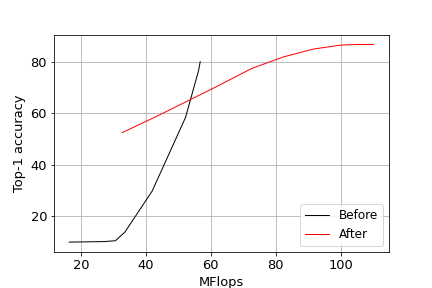}\label{approach2:d}}}
  
  \caption{The sparsity rates and exit performances for Approach 2.}
  \label{fig:approach2}
\end{figure}

In this approach, the ResNet-56 is pruned and fine tuned first. Then, linear layers are added to the model and these layers are pruned and fine tuned. The sparsity rate of each layer's weights and the exit performances after 20 iterations of pruning and fine tuning are shown in Fig.~\ref{fig:approach2}. We can make the following observations:

\begin{enumerate}
    \item Fig.~\ref{approach2:a} and Fig.~\ref{approach2:b} shows that although the final exit went through two pruning phases (one with base network, one with additional linear classifiers), it is not pruned much.
    
    \item Except the last exit, deeper exits are pruned more than earlier exits as seen from Fig.~\ref{approach2:a} and Fig.~\ref{approach2:b}.
    
    \item Compared to Fig.~\ref{fig:approach1}, linear layers become more sparse as seen from Fig.~\ref{approach2:b}.
    
    \item Unpruned baseline in Fig.~\ref{approach2:c} has high accuracy only at the last exit because earlier exits are attached after the base network was pruned and fine tuned. The additional linear classifiers were not trained at this point.
    
    \item Fine tuning linear layers pushes the computational cost up as seen in Fig.~\ref{approach2:d}, which is interesting.
\end{enumerate}

\subsection{Comparison}
\begin{figure}
  \centerline{\subfloat[The exit performances of the approaches.]{\includegraphics[width=0.46\linewidth]{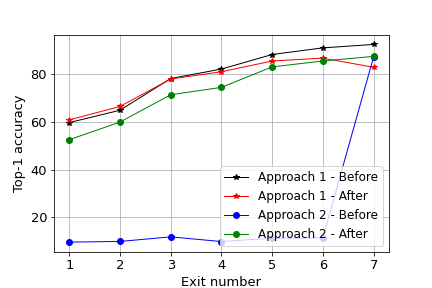}\label{comparison:a}}
  
  \subfloat[Accuracy-FLOPs plots of the approaches.]{\includegraphics[width=0.46\linewidth]{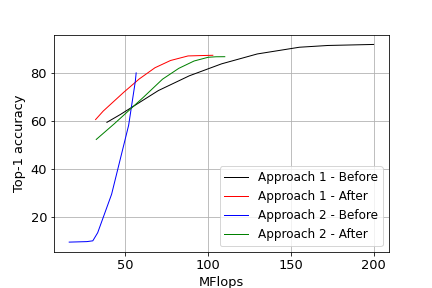}\label{comparison:b}}}
  
  \caption{Comparison of the approaches.}
  \label{fig:comparison}
\end{figure}

The exit performances of the two approaches are shown in Fig.~\ref{fig:comparison}. From Fig.~\ref{comparison:a}, it can be seen that that pruning all layers at the same time performs better at earlier exit points compared to pruning the layers in an ordered fashion. However, ordered pruning approach performs better at deeper exit points which give the best performance.

In terms of computational cost, pruning all layers at once reduces the computational cost by up to $15\%$ compared to ordered pruning, but the performances are close at high accuracy rates as seen from Fig.~\ref{comparison:b}. Both approaches are able to reduce the computational cost by half at the cost of $4\%$ decrease in the accuracy.

\section{Conclusion}
In this paper, we considered the problem of pruning early exit architectures. We evaluated the performance of two strategies in particular. First, intermediate classifiers are pruned jointly with the base network. Second, the base network is pruned first, followed by the intermediate classifiers. Although the former strategy outperforms the latter in general, the performance of the two strategies are close at high accuracy rates. Therefore, the processes of pruning and early exit can potentially be separated without significant penalty in performance. 

\begin{ack}
This work was supported in part by Army Research Lab (ARL) under Grant W911NF-21-2-0272, National Science Foundation (NSF) under Grant CNS-2148182, and by an award from the University of Illinois at Chicago Discovery Partners Institute Seed Funding Program.
\end{ack}

\bibliography{main}
\bibliographystyle{plainnat}

\end{document}